\documentclass[10.5pt,twocolumn,a4paper]{article}
\usepackage[utf8]{inputenc}
\usepackage{amsmath}
\usepackage{amsfonts}
\usepackage{amssymb}
\usepackage{multicol}
\usepackage{graphicx}
\usepackage{float}
\usepackage[none]{hyphenat}
\usepackage[framemethod=tikz]{mdframed}
\usepackage{rotating}

\usepackage[left=2cm,right=2cm,top=2cm,bottom=2cm]{geometry}

\usepackage{braket}
\usepackage{multirow}

\begin{document}

\newgeometry{left=2cm, right=2cm, top=1.5cm, bottom=2cm}

\title{{\LARGE Learning Multiple Categories on Deep Convolution Networks\\\begin{large}
Why deep convolution networks are effective in solving large recognition problems
\end{large}}}

\author{Mohamed  Hajaj  \hspace{1.5cm}   Duncan Gillies\\{\small Department of Computing, Imperial College London}}
\date{\vspace{-0.5ex}}
\maketitle

\newcommand*{\bfrac}[2]{\genfrac{}{}{0pt}{}{#1}{#2}}

\section*{Abstract}
\textit{Deep convolution networks have proved very successful with big datasets such as the 1000-classes ImageNet. Results show that the error rate increases slowly as the size of the dataset increases. Experiments presented here may explain why these networks are very effective in solving big recognition problems. If the big task is made up of multiple smaller tasks, then the results show the ability of deep convolution networks to decompose the complex task into a number of smaller tasks and to learn them simultaneously. The results show that the performance of solving the big task on a single network is very close to the average performance of solving each of the smaller tasks on a separate network. Experiments also show the advantage of using task specific or category labels in combination with class labels.}

\section{Introduction}

Since 2012 and starting with the introduction of Alex Krizhevsky model \cite{krizhevsky2012imagenet}, all the winners \cite{simonyan2014very, szegedy2015going, he2016deep, hu2017squeeze} of the classification competition part of the ImageNet challenge were deep convolution networks. The ImageNet dataset is made up of 1000-classes which is much bigger in size compared to earlier image recognition benchmarks such as the MNIST dataset, CIFAR10 dataset, CIFAR100 dataset etc. The experiments presented here investigate why these networks are very efficient in solving such big image recognition problems.

The first set of experiments measure how the performance of deep convolution networks changes as the number of classes increases. Multiple datasets with different sizes were randomly sampled from ImageNet, and the test error rate was measured at each size. For each size, multiple datasets were sampled and tested to reduce variance in the results. The results reveal that the error rate increases at a much lower rate compared to the increase in the number of classes. It is interesting to ask why the performance of these networks is resilient against increasing the number of classes.

The main experiments in this study may provide some insight into why deep convolution networks are very effective in solving large recognition problems. In this experiment a dataset made up of multiple categories (sampled from ImageNet) were used to train a deep convolution network, and the results were compared to the results of learning each small category on a separate network. The performance of the single network trained on all categories was very close to the average performance of the other networks trained on single categories. This means the network was able to break down the main task into smaller tasks and learn them simultaneously with very small drop in performance. The network has the remarkable inherent ability to recognize that a certain complex task (like ImageNet) is made up of multiple smaller tasks, and without any hint, is able to discover those smaller tasks and learn them simultaneously to solve the large main task.  

Finally, using the data available from the main experiment, an extra experiment showed that using both the class and category labels of the image outperformed the standard labeling scheme of only using the class labels. 

\section{Experiments}
\subsection{Performance vs number of classes}

The first part of this study tries to measure the performance of deep convolution networks in relation to the number of classes in the image dataset. Datasets with 5 different sizes were randomly sampled from the 1000-classes ImageNet. The number of classes of these datasets were 10, 50, 100, 500, and 1000 classes. To reduce variance in the results, multiple datasets were sampled at each size (except the last one) and the average performance is reported. The number of datasets sampled at each size were 10, 10, 5, 2, 1 respectively. The ImageNet dataset used here is the one used in the ILSVRC 2015 competition, and out of the 1000 classes, 891 of them had the maximum number of training images of 1300 images per class. All the classes sampled here belong to these 891, and therefore all the classes used in this experiment had 1300 training images (expect the one using the entire ImageNet of 1000 classes). The ImageNet validation set was used to sample the different test sets, and each class has 50 test images.  

\restoregeometry

Table (\ref{Table_NetworkStructure}) shows the structure of the 34-layer residual network used in this experiment. All datasets used the same structure with the only difference being the number of neurons in the output layer. Instead of using the standard data augmentation technique \cite{simonyan2014very} that is usually used with deep residual networks, the more aggressive augmentation method (usually used with the inception model \cite{szegedy2015going}) is used here. The reason for this choice is because it performs slightly better with smaller datasets (probably because it is more effective in reducing overfitting). The size of the cropped square is chosen randomly to be between 8\% and 100\% of the size of the maximum square in the image, and the aspect ratio is changed randomly to be between $3/4$ and $4/3$.
 
The method in \cite{he2015delving} was used to initialize the network weights, and the standard color augmentation method in \cite{krizhevsky2012imagenet} was used to simulate variance in illumination and intensity that exists in natural images. The RMSProp optimization method was used instead of gradient decent with momentum to update the network parameters, using a decay value of 0.999 to calculate the running average per parameter. RMSProp produces similar results to ADAM \cite{kingma2014adam} with the advantage of using a single running average per parameter instead of 2. 


\begin{table}[H]
\centering
\renewcommand{\arraystretch}{1.6}
\begin{tabular}{|c|c|}	
	\hline 
	
	$\mathrm{\bfrac{output}{size}}$ & 34 Layers \\ 
	
	\hline 
	
	{\scriptsize $112\times112$} & {\footnotesize $conv,\ 7\times 7$, stride 2\hspace{.12cm} 64} \\ 
	
	\hline 
	
	\multirow{2}{*}{\scriptsize {$56\times56$}} & {\footnotesize $max\ pool\ 3\times3$, stride 2}\\
	\cline{2-2}
	                              & {\small $\left[\bfrac{conv,\ 3\times3,\hspace{.2cm} 64}{conv,\ 3\times3,\hspace{.2cm} 64}\right]\times3$} \\[0.1cm]                             
	 
	\hline
	
	\multirow{2}{*}{{\scriptsize $28\times28$}} & {\footnotesize $max\ pool\ 3\times3$, stride 2}\\
	\cline{2-2}
	                              & {\small $\left[\bfrac{conv,\ 3\times3,\hspace{.2cm} 128}{conv,\ 3\times3,\hspace{.2cm} 128}\right]\times4$} \\[0.1cm]                             
	 
	\hline
	
	\multirow{2}{*}{{\scriptsize $14\times14$}} & {\footnotesize $max\ pool\ 3\times3$, stride 2}\\
	\cline{2-2}
	                              & {\small $\left[\bfrac{conv,\ 3\times3,\hspace{.2cm} 256}{conv,\ 3\times3,\hspace{.2cm} 256}\right]\times6$} \\[0.1cm]                             
	 
	\hline
	
	\multirow{2}{*}{{\scriptsize $7\times7$}} & {\footnotesize $max\ pool\ 3\times3$, stride 2}\\
	\cline{2-2}
	                              & {\small $\left[\bfrac{conv,\ 3\times3,\hspace{.2cm} 512}{conv,\ 3\times3,\hspace{.2cm} 512}\right]\times3$} \\[0.1cm]                             
	 
	\hline
	 	
	{\scriptsize $1\times1$} & {\small $\bfrac{global\ avg\ pool\ 7\times7}{10-\mathrm{d}\ fc,\ softmax}$}  \\[0.1cm]
	
	\hline                     
\end{tabular} 
\caption{{\small Network Structure}}
\label{Table_NetworkStructure}
\end{table}


Table (\ref{Table_Results_multiple_sizes}) shows the results for all dataset sizes. The results show the multi-crop error rate for all 5 sizes. As the number of classes increases by a factor of 10 from 10 to 100 to 1000 classes, the error rate only increases by a factor close to 2 from 4.81\% to 10.1\% to 21.2\%. This shows how effective these networks are in solving very large problems. The exact values of these results may vary based on the makeup of the datasets sampled from ImageNet, and to reduce this variance multiple datasets are sampled at each size. However, despite the small variance the error rate always grows at a much lower rate compared to the increase in the number of classes.   


\begin{table}[H]
\centering
\renewcommand{\arraystretch}{1.9}
\begin{tabular}{|c|c|c|c|c|c|}	
	\hline 
	
	{{\large $\mathrm{\bfrac{Data}{ Size}}$}} & $\mathrm{\bfrac{10}{ Classes}}$ & $\mathrm{\bfrac{50}{ Classes}}$ & $\mathrm{\bfrac{100}{ Classes}}$ & $\mathrm{\bfrac{500}{ Classes}}$ & $\mathrm{\bfrac{1000}{ Classes}}$\\ [0.2cm]
	
	\hline 
	
	{{\large $\mathrm{\bfrac{Test}{ Error}}$}} & 4.81\% & 7.7\% & 10.1\% & 16\% & 21.8\% \\[0.2cm]

	\hline                     
\end{tabular} 
\caption{{\small Results for datasets with different sizes sampled from ImageNet. }}
\label{Table_Results_multiple_sizes}
\end{table}

\begin{figure}[H]
\centering
\includegraphics[scale=0.8]{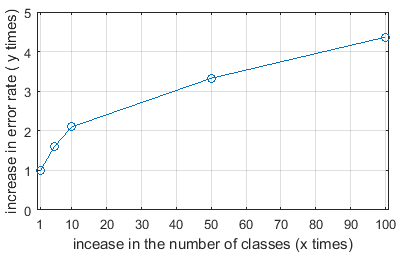} 
\caption{{\scriptsize the relative increase in error rate compared to the relative increase in the number of classes.}   }
\label{Curve_incease_in error_vs_increase_in classes}
\end{figure}

Figure (\ref{Curve_incease_in error_vs_increase_in classes}) shows the relative increase in the error rate compared to the relative increase in the number of classes. The relative increase in the error and in the number of classes are obtained by dividing all entries in table (\ref{Table_Results_multiple_sizes}) by the entries in the first column. As the number of classes is increased up 100 times, the error rate only increases 4.5 times.


\subsection{Classifying Multiple Categories }

As the previous experiment showed, one of the strengths of deep convolution networks is their ability to effectively solve large recognition problems such as the 1000-classes ImageNet. It also showed that the performance drops slowly as the size of the task increases significantly. Looking at the classes that make up the ImageNet dataset, there are many similar classes that can be divided into categories (e.g. multiple species of dogs, cats, birds, multiple types of cars etc.). It is often likely for very large datasets to contain similar classes that can be put into categories, and the next experiment tries to measure how deep convolution networks react to such similarities, and if that might explain the success of these networks with large datasets.


\begin{table*}
\centering
\renewcommand{\arraystretch}{2.2}
\begin{tabular}{|c|c|c|c|c|c|c|c|c|c|c|c|}	
	\hline 
	
	Category & \raisebox{-0.5em}{\rotatebox[origin=c]{90}{Cars}} & 
	\raisebox{-0.5em}{\rotatebox[origin=c]{90}{Bugs}}& 
	\raisebox{-0.5em}{\rotatebox[origin=c]{90}{Cats}}& 
	\raisebox{-0.5em}{\rotatebox[origin=c]{90}{China}} & 
	\raisebox{-0.5em}{\rotatebox[origin=c]{90}{Birds}}& 
	\raisebox{-0.5em}{\rotatebox[origin=c]{90}{Fruits}}&  
	\raisebox{-0.5em}{\rotatebox[origin=c]{90}{{\footnotesize Furniture}}}& 
	\raisebox{-0.5em}{\rotatebox[origin=c]{90}{Lizards}}& 
	\raisebox{-0.5em}{\rotatebox[origin=c]{90}{{\footnotesize Monkeys}}}& 
	\raisebox{-0.5em}{\rotatebox[origin=c]{90}{Snakes}} &
	\raisebox{-0.5em}{\rotatebox[origin=c]{90}{\textbf{Avg.}}}\\ [0.50cm]
	\hline 
	
	$\mathrm{\bfrac{Network\ Per}{Category}}$ & 16.6\% & 12.6\% & 19.0\% & 23.4\% & 2.43\% & 10.4\% & 16.4\% & 22.7\% & 23.4\% & 24.9\% & \textbf{17.18\%}\\ [0.125cm]
	
	\hline 
	
	{{\large $\mathrm{\bfrac{Shared}{ Network}}$}} & 17.4\% & 12.5\% & 20.4\% & 27.0\% & 2.37\% & 10.9\% & 18.9\% & 23.6\% & 24.5\% & 25.3\% & \textbf{18.28\%} \\[0.125cm]

	\hline                     
\end{tabular} 
\caption{{\small Results per category for the shared network vs the results obtained using separate network per category.}}
\label{Table_results_Per_Category}
\end{table*}


A dataset made up of 100 classes was constructed from ImageNet, where the chosen classes belong to 10 different categories. These categories are, birds, bugs, cars, cats, china and cookware, fruits and vegetables, furniture, lizards, monkeys, and snakes. Each of these naturally divided categories consists of 10 classes, for example the cars category is divided into ambulances, jeep (four wheel) cars, family cars, convertible cars, police cars, taxis, sports cars, small buses, large family cars, pickup cars. Each class has 1300 training images, and 50 test images (sampled from the ImageNet validation set). For reproducing the results, table (\ref{Folder_Names}) shows the folder names of the classes that make up all the categories. 

In the first part of this experiment all 10 categories are considered as a single dataset and used to train a deep convolution network. Images were labeled regularly using a vector of 100 numbers, with only one of them is ON to reflect a specific class. This is the regular way of coding image labels when softmax is used as the activation function of the output layer. Therefore, no hint is given to the network to treat these 100 classes as 10 separate categories.

In the second part of the experiment, each category was considered as a separate dataset, and used to train a separate network. Therefore, 10 separate networks will be trained using the 10 different categories, and the size and structure of these networks is the same as the size and structure of the shared network used to learn all categories. The only difference is the size of the output layer. 

The accuracy of classifying a category using a separate network will be compared to the accuracy of classifying that category on the shared network used to learn all categories. This comparison will measure the drop in performance per category for the shared network. The drop should reflect the added confusion caused by learning all the categories on the same network.

The same network structure shown in table (\ref{Table_NetworkStructure}) will be used here, with the same setup to the hyper-parameters used in the previous experiment. Table (\ref{Table_results_Per_Category}) shows the results per category for both parts of the experiment. The top row shows the results per category for the 10 separate networks, while the bottom row shows the results per category for the shared network. The last column shows the average results for all 10 categories. For most categories the results were very close with only a small drop in the performance of the shared network, as the average error increases from 17.18\% to 18.28\%.  

From the results in table (\ref{Table_results_Per_Category}), the shared network utilized the fact that the 100 classes belong to 10 different categories, and was able to learn all of them with accuracy very close to learning each one on a separate network. The network was able to break down the main task into multiple smaller tasks, and learn them simultaneously. Therefore, for a big task made up of multiple smaller tasks, what dictates the difficulty of learning the main task is not its size (number of classes), but rather the difficulty of learning each of the smaller tasks. From table (\ref{Table_results_Per_Category}), the performance of the network used to solve the main task was very close to the average performance of solving each of the subtasks separately.

If a big dataset is made of multiple groups where each group contains classes that are similar and hard to distinguish, then the difficulty of distinguishing between the members of these groups will probably decide the performance of the network. In reality however, big datasets such as ImageNet have a mixed bag of classes that can be separated into categories (cars, cats, dogs etc.), and classes that have common features with many other classes, and cannot be put into a specific group or category. Therefore, the performance of deep convolution networks on such big datasets, will be affected by both, the difficulty of the subtasks within the big dataset, as well as the size of the dataset. If most of the classes belong to well separated categories, then the difficulty of learning those categories will probably decide the performance of the network, while if most of the classes belong to a big vague group that cannot be broken down to smaller categories, then the size of the task will probably decide the performance of the network. The results of the previous experiment on datasets with different sizes sampled randomly from ImageNet shows something in between, where the performance of convolution networks drops slowly as the number of classes increases.

\begin{figure} [H]
\centering
\includegraphics[scale=0.6]{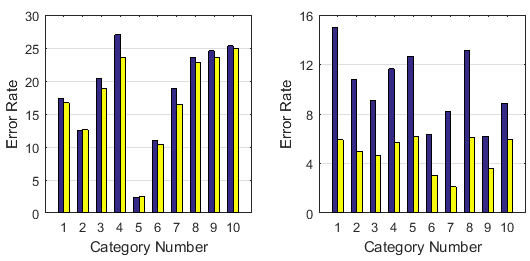} 
\caption{{\scriptsize The blue bars show the error rate per category obtained using the shared network, the yellow bars were obtained using a separate network per category. \textbf{left:-} for the naturally divided dataset, \textbf{right:-} for the randomly divided dataset.}   }
\label{Bars}
\end{figure}

In order to put these results into perspective, they will be compared with results obtained using another dataset with the same size, that is randomly sampled from ImageNet, and randomly divided into 10 groups. Figure (\ref{Bars}) shows the results per category (or group) for both cases. The yellow bars show the error rates per category (group) when each category is learned separately, and the blue bars show the error rates per category (group) for the shared network. The left figure shows the results for the naturally divided dataset, and the right figure shows the results for the randomly divided dataset. By visually comparing the two figures, we see that the shared network succeeded in learning all naturally divided groups, while it failed to do the same for the randomly divided groups.


\begin{table*}
\centering
\renewcommand{\arraystretch}{2.2}
\begin{tabular}{|c|c|c|c|c|c|c|c|c|c|c|c|}	
	\hline 
	
	Super-Class & \raisebox{-0.5em}{\rotatebox[origin=c]{90}{Cars}} & 
	\raisebox{-0.5em}{\rotatebox[origin=c]{90}{Bugs}}& 
	\raisebox{-0.5em}{\rotatebox[origin=c]{90}{Cats}}& 
	\raisebox{-0.5em}{\rotatebox[origin=c]{90}{China}} & 
	\raisebox{-0.5em}{\rotatebox[origin=c]{90}{Birds}}& 
	\raisebox{-0.5em}{\rotatebox[origin=c]{90}{Fruits}}&  
	\raisebox{-0.5em}{\rotatebox[origin=c]{90}{{\footnotesize Furniture}}}& 
	\raisebox{-0.5em}{\rotatebox[origin=c]{90}{Lizards}}& 
	\raisebox{-0.5em}{\rotatebox[origin=c]{90}{{\footnotesize Monkeys}}}& 
	\raisebox{-0.5em}{\rotatebox[origin=c]{90}{Snakes}} &
	\raisebox{-0.5em}{\rotatebox[origin=c]{90}{\textbf{Avg.}}}\\ [0.50cm]
	\hline 
	
	Error Rate & 0.56\% & 1.95\% & 1.86\% & 5.2\% & 0.34\% & 1.4\% & 2.86\% & 4.2\% & 0.83\% & 4.46\% & \textbf{2.36\%}\\ [0.125cm]

	\hline                     
\end{tabular} 
\caption{{\small results obtained by merging each category into a super-class. The results show the amount of leakage between categories.}}
\label{Table_leakage}
\end{table*}


\subsubsection{Inter-category Leakage }

Inter-category leakage measures the ratio of images in each category that have been misclassified as classes from other categories, when all categories are learned using a shared network. The leakage can be measured using the confusion matrix, by merging the results of all the classes that belong to a single category to form a single superclass that represents that category. The merger process goes like this: - if an image is misclassified as a class from the same category then this is considered a correct classification, while if an image is misclassified as a class from a different category then this is considered a wrong classification. When combining all the classes in each category into a single superclass, then the inter-category error rate or leakage is equal to 2.36\% as shown in table (\ref{Table_leakage}). The low inter-category leakage of 2.36\% shows that the network rarely misclassifies an image as one from a different category. Out of the 18.28\% misclassified images in table (\ref{Table_leakage}) only 2.36\% happened between categories, while 18.28 – 2.36\% = 15.92\% happened locally within each category.

\begin{figure} [H]
\centering
\includegraphics[scale=1.0]{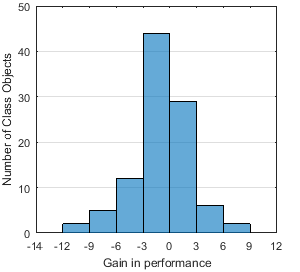} 
\caption{{\scriptsize Histogram of the difference in performance per class, between using the shared network, and using a separate network per category. 100 values for 100 classes. }   }
\label{Histogram}
\end{figure}

Table (\ref{Table_results_Per_Category}) showed a drop in performance for the shared network compared to using a separate network per category equal to 18.28 – 17.18 = 1.1\%, while table (\ref{Table_leakage}) shows a bigger leakage between categories for the shared network equal to 2.36\%. If 2.36\% of the images has suffered from being learned with other categories on the same network, then 2.36 - 1.1 = 1.26\% of the images must have benefited by being learned with other categories. Figure (\ref{Histogram}) explains this and shows a histogram for the difference in accuracy per class. It shows that about two third of the classes lost some performance by being learned on the shared network, while about one third has actually gained some performance. The shape of the histogram is close to a normal distribution with a negative average close to zero that reflects the 1.1\% drop in performance for the shared network, and a small variance that reflects the similarity between the results obtained using the shared network and the 10 separate networks. 


\subsection{Using Class and Category labels}

Using the dataset from the previous experiment (made up of 100 classes divided into 10 categories), a combined labeling scheme, that uses both class and category labels, will be tested against the standard labeling scheme that uses only class labels. The adopted method of adding the category label is straight forward and can easily be implemented. The combined class/category label is a vector made of 110 numbers, 10 numbers for each of the 10 categories, and 100 numbers for each of the 100 classes. For each image, 2 numbers will be ON, one represents the class of the image, and the other one represents the category of that image. In order to use a single softmax in the output layer, the two ON numbers per label are set to 0.5.

\vspace*{0.5cm}

\begin{table}[H]
\centering
\renewcommand{\arraystretch}{2.0}
\begin{tabular}{|c|c|}	
	\hline 
	
	Labels scheme & Error Rate \\ [0.125cm]
	\hline 
	
	class label & 18.28\%\\ [0.125cm]
	
	\hline 
	
	{\large $\mathrm{\bfrac{class/category}{label}}$} & 17.3\%\\[0.125cm]                  
	\hline
\end{tabular} 
\caption{{\small results using class/category labels vs using only class labels.}}
\label{Table_category_Label}
\end{table}

\begin{figure} [H]
\centering
\includegraphics[scale=0.4]{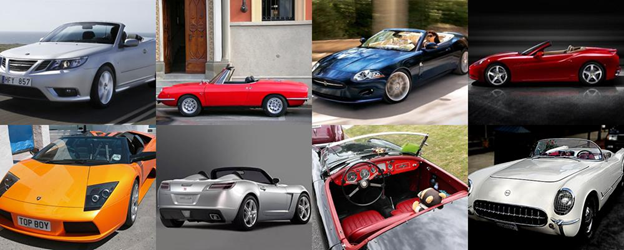} 
\caption{{\scriptsize convertible cars that were labeled as sports cars, In ImageNet. }   }
\label{cars}
\end{figure}

Table (\ref{Table_category_Label}) shows the results of using the combined class/category labeling scheme, vs using only the class label. There is about 1\% (about 5.3\% relative reduction) improvement in accuracy when the category label is added. This shows that the standard labeling method of using only the class label is very basic and can be improved. The construction of ImageNet is done by hand, and a closer inspection of the cars category shows some sports cars labeled as convertible cars and vice versa. This way of randomly labeling cars that share both attributes causes confusion to the network. If the image label was constructed using only the class number, then two convertible cars can have two completely different labels (one as a convertible car and the other as a sports car), while if the category number is added, then these two cars will at least share the same category number, and that will make their labels 50\% similar, rather than 0\% similar. Figure (\ref{cars}) shows some convertible cars from ImageNet that have been labeled as sports cars, because they share both attributes. In fact, both classes were among those that benefited from adding the category label, with the error rate for the convertible cars class dropping from 14.2\% to 10.2\%, and for the sports cars class dropping from 15.6\% to 12.8\%, which is much better than 1\% average improvement. 


\nocite{*}
\bibliographystyle{apalike}
\bibliography{References}

\begin{thebibliography}{}

\bibitem[Bengio et~al., 2009]{bengio2009learning}
Bengio, Y. et~al. (2009).
\newblock Learning deep architectures for ai.
\newblock {\em Foundations and trends{\textregistered} in Machine Learning},
  2(1):1--127.

\bibitem[Caruana, 1998]{caruana1998multitask}
Caruana, R. (1998).
\newblock Multitask learning.
\newblock In {\em Learning to learn}, pages 95--133. Springer.

\bibitem[Collobert and Weston, 2008]{collobert2008unified}
Collobert, R. and Weston, J. (2008).
\newblock A unified architecture for natural language processing: Deep neural
  networks with multitask learning.
\newblock In {\em Proceedings of the 25th international conference on Machine
  learning}, pages 160--167. ACM.

\bibitem[Goodfellow et~al., 2013]{goodfellow2013empirical}
Goodfellow, I.~J., Mirza, M., Xiao, D., Courville, A., and Bengio, Y. (2013).
\newblock An empirical investigation of catastrophic forgetting in
  gradient-based neural networks.
\newblock {\em arXiv preprint arXiv:1312.6211}.

\bibitem[He and Sun, 2015]{he2015convolutional}
He, K. and Sun, J. (2015).
\newblock Convolutional neural networks at constrained time cost.
\newblock In {\em Proceedings of the IEEE Conference on Computer Vision and
  Pattern Recognition}, pages 5353--5360.

\bibitem[He et~al., 2015]{he2015delving}
He, K., Zhang, X., Ren, S., and Sun, J. (2015).
\newblock Delving deep into rectifiers: Surpassing human-level performance on
  imagenet classification.
\newblock In {\em Proceedings of the IEEE international conference on computer
  vision}, pages 1026--1034.

\bibitem[He et~al., 2016a]{he2016deep}
He, K., Zhang, X., Ren, S., and Sun, J. (2016a).
\newblock Deep residual learning for image recognition.
\newblock In {\em Proceedings of the IEEE conference on computer vision and
  pattern recognition}, pages 770--778.

\bibitem[He et~al., 2016b]{he2016identity}
He, K., Zhang, X., Ren, S., and Sun, J. (2016b).
\newblock Identity mappings in deep residual networks.
\newblock In {\em European Conference on Computer Vision}, pages 630--645.
  Springer.

\bibitem[Hu et~al., 2017]{hu2017squeeze}
Hu, J., Shen, L., and Sun, G. (2017).
\newblock Squeeze-and-excitation networks.
\newblock {\em arXiv preprint arXiv:1709.01507}.

\bibitem[Huang et~al., 2016]{huang2016deep}
Huang, G., Sun, Y., Liu, Z., Sedra, D., and Weinberger, K.~Q. (2016).
\newblock Deep networks with stochastic depth.
\newblock In {\em European Conference on Computer Vision}, pages 646--661.
  Springer.

\bibitem[Ioffe and Szegedy, 2015]{ioffe2015batch}
Ioffe, S. and Szegedy, C. (2015).
\newblock Batch normalization: Accelerating deep network training by reducing
  internal covariate shift.
\newblock In {\em International Conference on Machine Learning}, pages
  448--456.

\bibitem[Kingma and Ba, 2014]{kingma2014adam}
Kingma, D. and Ba, J. (2014).
\newblock Adam: A method for stochastic optimization.
\newblock {\em arXiv preprint arXiv:1412.6980}.

\bibitem[Kirkpatrick et~al., 2017]{kirkpatrick2017overcoming}
Kirkpatrick, J., Pascanu, R., Rabinowitz, N., Veness, J., Desjardins, G., Rusu,
  A.~A., Milan, K., Quan, J., Ramalho, T., Grabska-Barwinska, A., et~al.
  (2017).
\newblock Overcoming catastrophic forgetting in neural networks.
\newblock {\em Proceedings of the National Academy of Sciences}, page
  201611835.

\bibitem[Krizhevsky et~al., 2012]{krizhevsky2012imagenet}
Krizhevsky, A., Sutskever, I., and Hinton, G.~E. (2012).
\newblock Imagenet classification with deep convolutional neural networks.
\newblock In {\em Advances in neural information processing systems}, pages
  1097--1105.

\bibitem[Liu et~al., 2015]{liu2015representation}
Liu, X., Gao, J., He, X., Deng, L., Duh, K., and Wang, Y.-Y. (2015).
\newblock Representation learning using multi-task deep neural networks for
  semantic classification and information retrieval.
\newblock In {\em HLT-NAACL}, pages 912--921.

\bibitem[Shimodaira, 2000]{shimodaira2000improving}
Shimodaira, H. (2000).
\newblock Improving predictive inference under covariate shift by weighting the
  log-likelihood function.
\newblock {\em Journal of statistical planning and inference}, 90(2):227--244.

\bibitem[Simonyan and Zisserman, 2014]{simonyan2014very}
Simonyan, K. and Zisserman, A. (2014).
\newblock Very deep convolutional networks for large-scale image recognition.
\newblock {\em arXiv preprint arXiv:1409.1556}.

\bibitem[Srivastava et~al., 2015]{srivastava2015highway}
Srivastava, R.~K., Greff, K., and Schmidhuber, J. (2015).
\newblock Highway networks.
\newblock {\em arXiv preprint arXiv:1505.00387}.

\bibitem[Szegedy et~al., 2017]{szegedy2017inception}
Szegedy, C., Ioffe, S., Vanhoucke, V., and Alemi, A.~A. (2017).
\newblock Inception-v4, inception-resnet and the impact of residual connections
  on learning.
\newblock In {\em AAAI}, pages 4278--4284.

\bibitem[Szegedy et~al., 2015]{szegedy2015going}
Szegedy, C., Liu, W., Jia, Y., Sermanet, P., Reed, S., Anguelov, D., Erhan, D.,
  Vanhoucke, V., and Rabinovich, A. (2015).
\newblock Going deeper with convolutions.
\newblock In {\em Proceedings of the IEEE conference on computer vision and
  pattern recognition}, pages 1--9.

\bibitem[Szegedy et~al., 2016]{szegedy2016rethinking}
Szegedy, C., Vanhoucke, V., Ioffe, S., Shlens, J., and Wojna, Z. (2016).
\newblock Rethinking the inception architecture for computer vision.
\newblock In {\em Proceedings of the IEEE Conference on Computer Vision and
  Pattern Recognition}, pages 2818--2826.

\bibitem[Zagoruyko and Komodakis, 2016]{zagoruyko2016wide}
Zagoruyko, S. and Komodakis, N. (2016).
\newblock Wide residual networks.
\newblock {\em arXiv preprint arXiv:1605.07146}.

\bibitem[Zhang et~al., 2014]{zhang2014facial}
Zhang, Z., Luo, P., Loy, C.~C., and Tang, X. (2014).
\newblock Facial landmark detection by deep multi-task learning.
\newblock In {\em European Conference on Computer Vision}, pages 94--108.
  Springer.

\end{thebibliography}

\newpage


\begin{sidewaystable*}
    \centering
    \renewcommand{\arraystretch}{2}
    \caption{{\small Folder names of 100 Classes sampled from ImageNet ILSVRC 2015 to form 10 categories, each category has 10 classes.}}
   \begin{tabular}{|c|c|c|c|c|c|c|c|c|c|c|}
   \hline
    Cars & n02701002 & n02814533 & n02930766 & n03100240 & n03594945 & n03769881 & n03770679 & n03930630 & n03977966 & n04285008\\
    \hline
   Bugs & n01773797 & n01774384 & n01775062 & n02165105 & n02167151 & n02168699 & n02174001 & n02177972 & n02229544 & n02233338\\
    \hline
    Cats & n02123045 & n02123159 & n02123394 & n02123597 & n02124075 & n02127052 & n02128385 & n02128757 & n02128925 & n02130308\\
     \hline
    China & n03063599 & n03063689 & n03443371 & n03775546 & n03786901 & n03950228 & n04398044 & n04522168 & n07920052 & n07930864\\
     \hline
    Birds & n01530575 & n01531178 & n01532829 & n01534433 & n01537544 & n01558993 & n01560419 & n01580077 & n01592084 & n01828970\\
    \hline
    Fruits & n07720875 & n07742313 & n07745940 & n07747607 & n07749582 & n07753113 & n07753592 & n07768694 & n12620546 & n12768682\\
    \hline
   Furniture & n02870880  & n03014705 & n03016953 & n03018349 & n03125729 & n03131574 & n03201208 & n03337140 & n04099969 & n04550184\\
    \hline
   Lizards & n01629819 & n01630670 & n01631663 & n01632458 & n01675722 & n01682714 & n01685808 & n01687978 & n01689811 & n01693334\\
    \hline
    Monkeys & n02483362 & n02486261 & n02486410 & n02487347 & n02488291 & n02492035 & n02492660 & n02493509 & n02493793 & n02494079\\
    \hline
    Snakes & n01728572 & n01728920 & n01729322 & n01734418 & n01735189 & n01740131 & n01742172 & n01753488 & n01755581 & n01756291\\
    \hline
    \end{tabular}
    \label{Folder_Names}
\end{sidewaystable*}


\end{document}